\documentclass[9pt,twoside]{extarticle}
\usepackage[papersize={17cm,24cm},margin=22mm,top=17mm,headsep=5mm]{geometry} 
\usepackage{amsmath}
\usepackage{amsthm}
\usepackage{amssymb}
\usepackage{enumerate}
\usepackage{verbatim} % verbatim input
\usepackage{titling} % customize title
\usepackage{csquotes} % context sensitive quotes
\usepackage{upquote} % allow correct quotes in verbatim
\usepackage[bottom]{footmisc} % put footnotes down at the bottom margin
\usepackage{graphicx} % \includegraphics
\usepackage[noend]{algpseudocode}
\usepackage[linesnumbered,ruled]{algorithm2e}
\usepackage{thmtools}
\usepackage{thm-restate}
\usepackage[dvipsnames]{xcolor}
\usepackage[colorlinks,linkcolor = Aquamarine, urlcolor  = violet, citecolor = YellowOrange, anchorcolor = violet]{hyperref}

\usepackage[english]{babel} % multilinguality
\date{}
\usepackage{natbib}
\setcitestyle{square,numbers,comma}
\renewenvironment{abstract}
  {\noindent\small %\quotation
  {\noindent{\large\textbf\abstractname. }%\par\nobreak\smallskip
  \thispagestyle{plain}
  }}
  {}
    
\usepackage{fancyhdr}
\fancypagestyle{plain}{% first page of chapter
  \fancyhead[C,R]{}
  \fancyfoot{} % comment to put page number at bottom
  }
  
\newcommand*\R[0]{\mathbb{R}}

\newcommand*\ddt[0]{\frac{d}{d t}}

\newcommand*\E[1]{\mathbb{E}\left[#1\right]}
\newcommand*\Ep[2]{\mathbb{E}_{#1}\left[#2\right]}

\renewcommand*\d[0]{\text{d}}

\newcommand\numberthis{\addtocounter{equation}{1}\tag{\theequation}}
\newcommand*\lrb[1]{\left[#1\right]}
\newcommand*\lrbb[1]{\left\{#1\right\}}
\newcommand*\lrp[1]{\left(#1\right)}
\newcommand*\lrn[1]{\left\|#1\right\|}
\newcommand*\eu[1]{\left\| #1\right\|}

\renewcommand*\d[0]{\delta}
\newcommand*\xt[0]{\tilde{x}}
\newcommand*\vt[0]{\tilde{v}}
\newcommand*\pt[0]{\tilde{p}}

\newcommand*\Phit[0]{\tilde{\Phi}}
\newcommand*\nh{\hat{\nabla}}
\newcommand*\xh[0]{\hat{x}}
\newcommand*\vh[0]{\hat{v}}
\newcommand*\ph[0]{\hat{p}}

\newcommand*\Phih[0]{\hat{\Phi}}
\newcommand*\wt[0]{\tilde{w}}
\newcommand*\wh[0]{\hat{w}}
\newcommand*\cvec[2]{\begin{bmatrix} #1\\#2\end{bmatrix}}
\def\lv{\lVert}
\def\rv{\rVert}
\def\ke{\mathcal{E}_K}
\newcommand{\BlackBox}{\rule{1.5ex}{1.5ex}}  % end of proof
\newenvironment{Proof}{\par\noindent{\bf Proof\ }}{\hfill\BlackBox\\[2mm]}
\newtheorem{theorem}{Theorem}

\newtheorem{lemma}[theorem]{Lemma}

\newtheorem{remark}[theorem]{Remark}
\newtheorem{corollary}[theorem]{Corollary}

\newcommand*\samethanks[1][\value{footnote}]{\footnotemark[#1]}

\title{Underdamped Langevin MCMC: A non-asymptotic analysis} % do not capitalize every open class word
\author{Xiang Cheng  \thanks{X. Cheng and N. S. Chatterji contributed equally to this work.} \thanks{x.cheng@berkeley.edu; Computer Science Division, UC Berkeley; work performed while at Adobe Research.} \and 
  Niladri S. Chatterji \samethanks[1] \thanks{niladri.chatterji@berkeley.edu; Department of Physics, UC Berkeley.} \and Peter L. Bartlett \thanks{peter@berkeley.edu; Computer Science Division \& Department of Statistics, UC Berkeley, and Mathematical Sciences, QUT.} \and Michael I. Jordan \thanks{jordan@cs.berkeley.edu; Computer Science Division \& Department of Statistics, UC Berkeley.}}
\date{\today}
\begin{document}
\maketitle

\begin{abstract}
We study the underdamped Langevin diffusion when the log of the target distribution is smooth and strongly concave. We present a MCMC algorithm based on its discretization and show that it achieves $\varepsilon$ error (in 2-Wasserstein distance) in $\mathcal{O}(\sqrt{d}/\varepsilon)$ steps. This is a significant improvement over the best known rate for overdamped Langevin MCMC, which is $\mathcal{O}(d/\varepsilon^2)$ steps under the same smoothness/concavity assumptions.

The underdamped Langevin MCMC scheme can be viewed as a version of Hamiltonian Monte Carlo (HMC) which has been observed to outperform overdamped Langevin MCMC methods in a number of application areas. We provide quantitative rates that support this empirical wisdom.
\end{abstract}
\section{Introduction}
%In this document, study an algorithm for sampling from the target distribution $p^*(x,v) \propto e^{-\beta(f(x) - \frac{1}{2u}\|v\|_2^2)}$.
%
%The SDE \eqref{e:exactlangevindiffusion} has $p^*$ as its stationary distribution. We establish in section \ref{s:exactconvergence} that the the distributions $p_t$ from \eqref{e:exactlangevindiffusion} converges to $p^*$ geometrically in $W_2$.
%
%However, \eqref{e:exactlangevindiffusion} is a continuous time process and is not implementable. We will approximate \eqref{e:exactlangevindiffusion} by \eqref{e:discretelangevindiffusion}. We study the discretization error of \eqref{e:discretelangevindiffusion} in section \ref{s:discreteerror}.
%
%In section \ref{s:discreteconvergence}, we combine the results of sections \ref{s:exactconvergence} and \ref{s:discreteerror} to obtain our main convergence result, Theorem \ref{t:kstepconvergence}.
%
%Finally, in section \ref{s:howtodiscretize}, we give explicit formulas for how to implement \ref{e:discretelangevindiffusion}.
%
%
%\begin{remark}
%The $\frac{1}{L}$ factor is not an issue because we can assume wlog that $L=1$ by scaling the space $R^d$ by a factor of $L$.
%\end{remark} \blfootnote{(x.cheng, niladri.chatterji, peter)@berkeley.edu, jordan@cs.berkeley.edu}
In this paper, we study the continuous time \emph{underdamped} Langevin diffusion represented by the following stochastic differential equation (SDE):
\begin{align}\label{e:exactlangevindiffusion}
d v_t &= -\gamma v_t dt - u \nabla  f(x_t) dt + (\sqrt{2\gamma u}) dB_t \\
\nonumber d x_t &= v_t dt,
\end{align}
where $(x_t,v_t)\in \mathbb{R}^{2d}$, $f$ is a twice continuously-differentiable function and $B_t$ represents standard Brownian motion in $\mathbb{R}^d$. Under fairly mild conditions, it can be shown that the invariant distribution of the continuous-time process \eqref{e:exactlangevindiffusion} is proportional to $\exp(-(f(x)+ \lVert  v \rVert_2^2/2u))$. Thus the marginal distribution of $x$ is proportional to $\exp(-f(x))$. There is a discretized version of \eqref{e:exactlangevindiffusion} which can be implemented algorithmically, and provides a useful way to sample from $p^*(x) \propto e^{-f(x)}$ when the normalization constant is not known.

Our main result establishes the convergence of \eqref{e:exactlangevindiffusion} as well as its discretization, to the invariant distribution. This provides explicit rates for sampling from log-smooth and strongly log-concave distributions using the underdamped Langevin MCMC algorithm (Algorithm \ref{ulmcmc}).  

Underdamped Langevin diffusion is particularly interesting because it contains a Hamiltonian component, and its discretization can be viewed as a form of Hamiltonian MCMC. Hamiltonian MCMC (see review of HMC in \citep{neal,betan}) has been empirically observed to converge faster to the invariant distribution compared to standard Langevin MCMC which is a discretization of \emph{overdamped} Langevin diffusion,
\begin{align*}
dx_t = -\nabla f(x_t) dt + \sqrt{2}dB_t
\end{align*}
the first order SDE corresponding to the high friction limit of \eqref{e:exactlangevindiffusion}. This paper provides a non-asymptotic quantitative explanation for this statement.
\subsection{Related Work}
The first explicit proof of non-asymptotic convergence of overdamped Langevin MCMC for  log-smooth and strongly log-concave distributions was given by \citep{dalalyan}, where it was shown that discrete, overdamped Langevin diffusion achieves $\varepsilon$ error, in total variation distance, in $\mathcal{O}\left(\frac{d}{\varepsilon^2}\right)$ steps. Following this, \citep{durmus} proved that the same algorithm achieves $\varepsilon$ error, in 2-Wasserstein distance, in $\mathcal{O}\left(\frac{d}{\varepsilon^2}\right)$ steps. \citep{cheng2017convergence} obtained results similar to those in \citep{dalalyan} when the error is measured by KL-divergence. Recently \citep{raginsky2017non,dalalyan2017user} also analyzed convergence of overdamped Langevin MCMC with stochastic gradient updates. Asymptotic guarantees for overdamped Langevin MCMC was established much earlier in \citep{gelfand1991recursive,roberts1996exponential}.

Hamiltonian Monte Carlo (HMC) is a broad class of algorithms which involve Hamiltonian dynamics in some form. We refer to \citep{ma} for a survey of the results in this area. Among these, the variant studied in this paper (Algorithm \ref{ulmcmc}), based on  the discretization of \eqref{e:exactlangevindiffusion}, has a natural physical interpretation as the evolution of a particle's dynamics under a force field and drag. This equation was first proposed by \citep{kramers1940brownian} in the context of chemical reactions. The continuous-time process has been studied extensively~\citep{herau2002isotropic,dric2009hypocoercivity,eberle2017couplings,gorham2016measuring,baudoin2016wasserstein,bolley2010trend, calogero2012exponential, dolbeault2015hypocoercivity, mischler2014exponential}.

However, to the best of our knowledge there has been no prior polynomial-in-dimension convergence result for any version of HMC under a log-smooth or strongly log-concave assumption for the target distribution\footnote{Following the first version of this paper, two recent papers also independently analyzed and provided non-asymptotic guarantees for different versions of HMC \citep{mangoubi2017rapid,lee2017convergence} }. Most closely related to our work is the recent paper \cite{eberle2017couplings} that demonstrated a contraction property of the continuous-time process \eqref{e:exactlangevindiffusion}. That result deals, however, with a much larger class of functions, and because of this the distance to the invariant distribution scales exponentially with dimension $d$.

Also related is the recent work on understanding acceleration of first-order optimization methods as discretizations of second-order differential equations \citep{su2014differential,krichene2015accelerated, wibisono2016variational}.
\subsection{Contributions}
Our main contribution in this paper is to prove that Algorithm \ref{ulmcmc}, a variant of HMC algorithm, converges to $\varepsilon$ error in 2-Wasserstein distance after $\mathcal{O}\left(\frac{\sqrt{d}\kappa^2}{\varepsilon}\right)$ iterations, under the assumption that the target distribution is of the form $p^*\propto \exp(-(f(x))$, where $f$ is $L$ smooth and $m$ strongly convex (see section \ref{ss:assumptions}), with $\kappa= L/m$ denoting the condition number. Compared to the results of \cite{durmus} on the convergence of Langevin MCMC in $W_2$ in $\mathcal{O}\left(\frac{d\kappa^2}{\varepsilon^2}\right)$ iterations, this is an improvement in both $d$ and $\epsilon$. We also analyze the convergence of chain when we have noisy gradients with bounded variance and establish non-asymptotic convergence guarantees in this setting.
\subsection{Organization of the Paper}
In the next subsection we establish the notation and assumptions that we use throughout the paper. In Section \ref{s:resultsmain} we present the discretized version of \eqref{e:exactlangevindiffusion} and state our main results for convergence to the invariant distribution. Section \ref{s:exactconvergence} then establishes exponential convergence for the continuous-time process and in Section \ref{s:discreteerror} we show how to control the discretization error. Finally in Section \ref{s:discreteconvergence} we prove the convergence of the discretization of \eqref{e:exactlangevindiffusion}. We defer technical lemmas to the appendix.
%\subsection{Assumptions}\label{ss:assumptions}
%\input{assump}
\subsection{Notation and Definitions}
\label{s:definitions}
In this section, we present basic definitions and notational conventions. Throughout, we let
$\lv v \rv_{2}$ denotes the Euclidean norm, for a vector $v \in \mathbb{R}^{d}$.

\subsubsection{Assumption on $f$}\label{ss:assumptions}
We make the following assumptions regarding the function $f$.

\begin{enumerate}[({A}1)]
\item The function $f$ is twice continuously-differentiable on $\mathbb{R}^d$ and has Lipschitz continuous gradients; that is, there exists a positive constant $L >0$ such that for all $x,y \in \mathbb{R}^{d}$ we have
\begin{align*}
\lVert \nabla f(x) - \nabla f(y) \rVert_2 \le L \lVert x-y \rVert_2.
\end{align*}
\item $f$ is $m$-strongly convex, that is, there exists a positive constant $m>0$ such that for all $x,y \in \mathbb{R}^d$,
\begin{align*}
f(y) \ge f(x) + \langle \nabla f(x),y-x \rangle + \frac{m}{2} \lVert x-y \rVert_2^2.
\end{align*}
\end{enumerate}

It is fairly easy to show that under these two assumptions the Hessian of $f$ is positive definite throughout its domain, with $m I_{d\times d}  \preceq  \nabla^2 f(x) \preceq L I_{d\times d}$. We define $\kappa = L/m$ as the condition number. Throughout the paper we denote the minimum of $f(x)$ by $x^*$. Finally, we assume that we have a gradient oracle $\nabla f(\cdot)$; that is, we have access to $\nabla f(x)$ for all $x$.

\subsubsection{Coupling and Wasserstein Distance}
Denote by $\mathcal{B}(\mathbb{R}^d)$ the Borel $\sigma$-field of $\mathbb{R}^d$. Given probability measures $\mu$ and $\nu$ on $(\mathbb{R}^d,\mathcal{B}(\mathbb{R}^d))$, we define a \emph{transference plan} $\zeta$ between $\mu$ and $\nu$ as a probability measure on $(\mathbb{R}^d \times \mathbb{R}^d,\mathcal{B}(\mathbb{R}^d\times \mathbb{R}^d))$ such that for all sets $A \in \mathbb{R}^d$, $\zeta(A\times \mathbb{R}^d) = \mu(A)$ and $\zeta( \mathbb{R}^d \times A) = \nu(A)$. We denote $\Gamma(\mu,\nu)$ as the set of all transference plans. A pair of random variables $(X,Y)$ is called a coupling if there exists a $\zeta \in \Gamma(\mu,\nu)$ such that $(X,Y)$ are distributed according to $\zeta$. (With some abuse of notation, we will also refer to $\zeta$ as the coupling.)

We define the Wasserstein distance of order two between a pair of probability measures as follows:
\begin{align*}
W_2(\mu,\nu) \triangleq \left(\inf_{\zeta\in\Gamma(\mu,\nu)} \int \lv x-y\rv_2^2 d\zeta(x,y) \right)^{1/2}.
\end{align*}
Finally we denote by $\Gamma_{opt}(\mu,\nu)$ the set of transference plans that achieve the infimum in the definition of the Wasserstein distance between $\mu$ and $\nu$ (for more properties of $W_2(\cdot,\cdot)$ see \citep{villani}). 

\subsubsection{Underdamped Langevin Diffusion} \label{ss:underdampedlangevindiffusionnotation}
Throughout the paper we use $B_t$ to denote standard Brownian motion~\citep{brownian}. Next we set up the notation specific to the continuous and discrete processes that we study in this paper.
\begin{enumerate}
%\item Let $f: \R^d \to R$ be a $m$ strongly convex and $L$ smooth function, and let $\kappa = m/L$ be the inverse of the condition number. 
%\item Given two distributions $p$ and $p'$, we denote by $\Gamma(p,p')$ the set of all couplings between $p$ and $p'$ (a coupling is a joint distribution whose marginals match its two arguments). 
\item Consider the exact underdamped Langevin diffusion defined by the SDE \eqref{e:exactlangevindiffusion}, with an initial condition $(x_0,v_0)\sim p_0$ for some distribution $p_0$ on $\R^{2d}$. Let $p_t$ denote the distribution of $(x_t,v_t)$ and let $\Phi_t$ denote the operator that maps from $p_0$ to $p_t$:
\begin{equation}\label{d:phi}
\Phi_t  p_0 = p_t.
\end{equation}
%\begin{enumerate}
%%\item Let $\Phi_s$ denote the (stochastic) map that maps a point $(x_0,v_0)$ to $(x_t,v_t)$ under \eqref{e:exactlangevindiffusion}.
%\item We let $p_t(x,v)$ denote the distribution of $(x_t, v_t)$ with an initial condition of the form $(x_0,v_0)\sim p_0$ for some given $p_0$ must be specified for $p_t$ to be defined.
%\item  We denote by 
%\begin{equation}\label{d:phi}
%\Phi_t  p_0
%\end{equation}
%the distribution $p_t$ of $(x_t,v_t)$ given the initial condition $ (x_0,v_0) \sim p_0$.
%%\item It can be shown that the invariant distribution $p_\infty(x,v)$ for \eqref{e:exactlangevindiffusion} is the desired stationary distribution $p^*(x,v) \propto e^{-(f(x) + \frac{1}{2u}\|v\|_2^2)}$, under mild conditions on $f(\cdot)$.
%\end{enumerate}
\item One step of the discrete underdamped Langevin diffusion is defined by the SDE
\begin{align}
\label{e:discretelangevindiffusion}
d\vt_t &= -\gamma \vt_t dt -u \nabla f(\xt_0) dt + (\sqrt{2\gamma u}) dB_t \\
d \xt_t &= \vt_s dt, \nonumber
\end{align}
with an initial condition $(\xt_0,\vt_0)\sim \pt_0$.
Let $\pt_t$ and $\Phit_t$ be defined analogously to $p_t$ and $\Phi_t$ for $(x_t,v_t)$.

\textbf{Note 1:} The discrete update differs from \eqref{e:exactlangevindiffusion} by using $\xt_0$ instead of $\xt_t$ in the drift of $\vt_s$.

\textbf{Note 2:} We will only be analyzing the solutions to \eqref{e:discretelangevindiffusion} for small $t$. Think of an integral solution of \eqref{e:discretelangevindiffusion} as a single step of the discrete Langevin MCMC.
\end{enumerate}
\subsubsection{Stationary Distributions}
Throughout the paper, we denote by $p^*$ the unique distribution which satisfies $p^*(x,v) \propto \exp{-(f(x)+\frac{1}{2u}\lv v\rv_2^2)}$.
It can be shown that $p^*$ is the unique invariant distribution of \eqref{e:exactlangevindiffusion} (see, for example, Proposition 6.1 in \citep{pav}).

 Let $g(x,v) = (x,x+v)$. We let $q^*$ be the distribution of $g(x,v)$ when $(x,v) \sim p^*$.

\section{Results}
\label{s:resultsmain}
\subsection{Algorithm}
The underdamped Langevin MCMC algorithm that we analyze in this paper in shown in Algorithm~\ref{ulmcmc}.

\label{algosec}\begin{algorithm}[t] \label{ulmcmc}
\caption{Underdamped Langevin MCMC} 
\SetKwInOut{Input}{Input}
    \SetKwInOut{Output}{Output}
        \Input{Step size $\delta<1$, number of iterations $n$, initial point $(x^{(0)},0)$, smoothness parameter $L$ and gradient oracle $\nabla f(\cdot)$}

    \For {$i=0,1,\ldots,n-1$} 
      {
		  Sample $(x^{i+1},v^{i+1})\sim Z^{i+1}(x^i,v^i)$
        }
   \end{algorithm} 
The random vector $Z^{i+1}(x_i,v_i)\in \R^{2d}$, conditioned on $(x^i,v^i)$, has a Gaussian distribution with conditional mean and covariance obtained from the following computations:
   \begin{align*}
&\E{v^{i+1}} = v^i e^{-2 \d} - \frac{1}{2L}(1-e^{-2 \d}) \nabla f(x^i)\\
&\E{x^{i+1}}  = x^i + \frac{1}{2}(1-e^{-2 \d})v^i - \frac{1}{2L} \left( \d - \frac{1}{2}\left(1-e^{-2 \d}\right) \right) \nabla f(x^i)\\
&\E{\left(x^{i+1} - \E{x^{i+1}}\right) \left(x^{i+1} - \E{x^{i+1}}\right)^{\top}}= \frac{1}{L } \left[\d-\frac{1}{4}e^{-4\d}-\frac{3}{4}+e^{-2\d}\right] \cdot I_{d\times d}\\
&\E{\left(v^{i+1} - \E{v^{i+1}}\right) \left(v^{i+1} - \E{v^{i+1}}\right)^{\top}} = \frac{1}{L}(1-e^{-4 \d})\cdot I_{d\times d}\\
&\E{\left(x^{i+1} - \E{x^{i+1}}\right) \left(v^{i+1} - \E{v^{i+1}}\right)^{\top}}= \frac{1}{2L} \left[1+e^{-4\d}-2e^{-2\d}\right] \cdot I_{d \times d}.
\end{align*}

The distribution is obtained by integrating the discrete underdamped Langevin diffusion \eqref{e:discretelangevindiffusion} up to time $\delta$, with the specific choice of $\gamma=2$ and $u=1/L$. In other words, if $p^{(i)}$ is the distribution of $(x^i,v^i)$, then $Z^{i+1}(x^i,v^i) \sim p^{(i+1)} = \Phit_{\d} p^{(i)}$. Refer to Lemma \ref{l:gaussianexpressionforsamplingxtvt} in Appendix \ref{s:howtodiscretize} for the derivation.

\subsection{Main Result}
\begin{theorem} \label{t:kstepconvergence}
Let $p^{(n)}$ be the distribution of the iterate of Algorithm \ref{ulmcmc} after $n$ steps starting with the initial distribution $p^{(0)}(x,v) =1_{x=x^{(0)}} \cdot 1_{v=0}$. Let the initial distance to optimum satisfy $  \lv x^{(0)} - x^* \rv_2^2 \le \mathcal{D}^2$. If we set the step size to be  
$$\d = \frac{\varepsilon }{104\kappa} \sqrt{\frac{1}{d/m + \mathcal{D}^2}}, $$
and run Algorithm \ref{ulmcmc} for $n$ iterations with 
$$n \ge  \left( \frac{52\kappa^2 }{\varepsilon}\right)\cdot \left( \sqrt{ \frac{d}{m}+ \mathcal{D}^2}\right) \cdot \log\left( \frac{24\left(\frac{d}{m}+ \mathcal{D}^2 \right)}{\varepsilon}\right),$$
then we have the guarantee that 
\begin{align*}
W_2(p^{(n)} , p^*)\leq \varepsilon.
\end{align*}
\end{theorem}
\begin{remark}\label{r:comparetolangevin}
The dependence of the runtime on $d,\varepsilon$ is thus $\tilde{\mathcal{O}}\left(\frac{\sqrt{d}}{\varepsilon}\right)$, which is a significant improvement over the corresponding $\mathcal{O}\left(\frac{d}{\varepsilon^2}\right)$ runtime of (overdamped) Langevin diffusion in \cite{durmus}. 
\end{remark}
We note that the $\log(24(d/m + \mathcal{D}^2)/\varepsilon)$ factor can be shaved off by using a time-varying step size. We present this result as Theorem \ref{t:loggone} in Appendix \ref{s:stepsizechange}. In neither theorem have we attempted to optimize the constants. 
\subsubsection{Result with Stochastic Gradients}\label{ss:stochasticgradient}
Now we state convergence guarantees when we have access to noisy gradients, $\nh f(x) = \nabla f(x) + \xi$, where $\xi$ is a independent random variable that satisfies
%stochastic gradients at each step. We assume that the following about the  stochasticity
\begin{enumerate}
\item The noise is unbiased -- $\E{\xi}=0$.

\item The noise has bounded variance -- $\mathbb{E}[\|\xi\|_2^2]\leq d\sigma^2.$
\end{enumerate} 
Each step of the dynamics is now driven by the SDE,
\begin{align}
\label{e:discretelangevindiffusionwithnoise}
d\vh_t &= -\gamma \vh_t dt -u \nh f(\xh_0) dt + (\sqrt{2\gamma u}) dB_t \\
d \xh_t &= \vh_s dt, \nonumber
\end{align}
with an initial condition $(\xh_0,\vh_0)\sim \ph_0$.
Let $\ph_t$ and $\Phih_t$ be defined analogously to $p_t$ and $\Phi_t$ for $(x_t,v_t)$ in Section \ref{ss:underdampedlangevindiffusionnotation}.
\begin{theorem}[Proved in Appendix \ref{app:stochastic}] \label{t:stochasticconvergence}
 Let $p^{(n)}$ be the distribution of the iterate of Algorithm \ref{alg:ulmcmcnoise} (stated in Appendix \ref{app:stochastic}) after $n$ steps starting with the initial distribution $p^{(0)}(x,v) =1_{x=x^{(0)}} \cdot 1_{v=0}$. Let the initial distance to optimum satisfy $  \lv x^{(0)} - x^* \rv_2^2 \le \mathcal{D}^2$. If we set the step size to be  
$$\d = \min\lrbb{\frac{\varepsilon }{\kappa} \sqrt{\frac{5}{479232\left(d/m + \mathcal{D}^2\right)}}, \frac{\varepsilon^2L^2}{1440 \sigma^2 d\kappa}}, $$
and run Algorithm \ref{ulmcmc} for $n$ iterations with 
$$n \ge  \frac{\kappa}{\delta} \cdot \log\left( \frac{36\left(\frac{d}{m}+\mathcal{D}^2 \right)}{\varepsilon}\right),$$
then we have the guarantee that 
\begin{align*}
W_2(p^{(n)} , p^*)\leq \varepsilon.
\end{align*}
\end{theorem}
\begin{remark} Note that when the variance in the gradients -- $\sigma^2d$ is large we recover back the rate of overdamped Langevin diffusion and we need $\tilde{\mathcal{O}}(\sigma^2 \kappa^2 d /\epsilon^2)$ steps to achieve accuracy of $\varepsilon$ in $W_2$. 
\end{remark}
\section{Convergence of the Continuous-Time Process}
\label{s:exactconvergence}
In this section we prove Theorem \ref{t:contraction}, which demonstrates a contraction for solutions of the SDE \eqref{e:exactlangevindiffusion}. We will use Theorem \ref{t:contraction} along with a bound on the discretization error between \eqref{e:exactlangevindiffusion} and \eqref{e:discretelangevindiffusion} to establish guarantees for Algorithm \ref{ulmcmc}.

\begin{theorem}\label{t:contraction}
Let $(x_0,v_0)$ and $(y_0,w_0)$ be two arbitrary points in $\R^{2d}$.
Let $p_0$ be the Dirac delta distribution at $(x_0,v_0)$ and let $p_0'$ be the Dirac delta distribution at $(y_0,w_0)$. We pick $u = 1/L$ where $L$ is the smoothness parameter of the function $f(x)$ and $\gamma = 2$. Then for every $t>0$, there exists a coupling $\zeta_t(x_0,v_0,y_0,w_0) \in \Gamma(\Phi_t p_0,\Phi_t p_0')$ such that 
\begin{align}\label{e:conttimeconv}
& \Ep{(x_t,v_t,y_t,w_t) \sim \zeta_t((x_0,v_0,y_0,w_0))}{\|x_t - y_t\|_2^2 + \|(x_t+v_t)-(y_t+w_t)\|_2^2} \\
\nonumber &\qquad \qquad \qquad \qquad \leq e^{-t/\kappa}\left\{\|x_0 - y_0\|_2^2 + \|(x_0+v_0)-(y_0+w_0)\|_2^2\right\}.
\end{align}  
\end{theorem}
\begin{remark}
A similar objective function was used in \cite{eberle2017couplings} to prove contraction.
\end{remark}
Given this theorem it is fairly easy to establish the exponential convergence of the continuous-time process to the stationary distribution in $W_2$.
\begin{corollary}\label{c:exactconvergence}
Let $p_0$ be arbitrary distribution with $(x_0,v_0) \sim p_0$.  Let $q_0$ and $\Phi_t q_0$ be the distributions of  $(x_0,x_0 + v_0)$ and $(x_t,x_t+v_t)$, respectively (i.e., the images of $p_0$ and $\Phi_t p_0$ under the map $g(x,v) = (x,x+v)$). Then
$$W_2(\Phi_t q_0,q^*) \leq e^{- t/2\kappa} W_2(q_0,q^*).$$
\end{corollary}
\begin{Proof} We let $\zeta_0 \in \Gamma(p_0,p^*)$ such that $\mathbb{E}_{\zeta_0}\left[\lv x_0 - y_0 \rv_2^2 + \| x_0 - y_0 + v_0 -w_0 \|_2^2 \right] = W_2^2(q_0,q^*)$. %We denote by $\zeta_t$ a coupling between $(x_t,v_t)$ and $(y_t,w_t)$. Specifically 
For every $x_0,v_0,y_0,w_0$ we let $\zeta_t(x_0,v_0,y_0,w_0)$ be the coupling as prescribed by Theorem \ref{t:contraction}. Then we have,%We denote by $\zeta_t$ a coupling between $(x_t,v_t)$ and $(y_t,w_t)$. Specifically we defined $\zeta_t$ point-wise, to be a coupling such that conditioned on $(x_0,y_0,v_0,w_0)$, $\zeta_t$ is the coupling required by Theorem \ref{t:contraction} (we know that it exists).
\begin{align*}
& W_2^2(q_t,q^*)\\ 
& \overset{(i)}{\le} \mathbb{E}_{(x_0,v_0,y_0,w_0)\sim\zeta_0}\left[\mathbb{E}_{(x_t,v_t,y_t,w_t)\sim\zeta_t(x_0,v_0,y_0,w_0)}\left[ \lv x_t - y_t \rv_2^2 + \lv x_t - y_t+v_t-w_t\rv_2^2 \Big\lvert x_0,y_0,v_0,w_0   \right] \right] \\
& \overset{(ii)}{\le}  \mathbb{E}_{(x_0,v_0,y_0,w_0)\sim\zeta_0}\left[e^{-t/\kappa}\lrp{\lv x_0 - y_0 \rv_2^2 + \lv x_0 - y_0+v_0-w_0\rv_2^2 }\right] \\
& \overset{(iii)}{=} e^{-t/\kappa}W^2_2(q_0,q^*),
\end{align*}
where $(i)$ follows as the Wasserstein distance is defined by the optimal coupling and by the tower property of expectation, $(ii)$ follows by applying Theorem \ref{t:contraction} and finally $(iii)$ follows by choice of $\zeta_0$ to be the optimal coupling. One can verify that the random variables $(x_t, x_t+v_t, y_t, y_t+w_t)$ $(i)$ defines a valid coupling between $q_t$ and $q^*$. Taking square roots completes the proof.
%We let $\zeta_0 \in \Gamma_{opt} (q_0,q^*)$ (there is only one valid coupling as $q_0$ is a Dirac delta distribution).
%\begin{align*}
%\lefteqn{W_2(\Phi_t q_0, q^*)} \\
%& \leq \Ep{(x_t,v_t,y_t,w_t) \sim \zeta_t}{\|x_t - y_t\|_2^2 + \|(x_t+v_t)-(y_t+w_t)\|_2^2} \\
%& \leq \Ep{(x_0,v_0,y_0,w_0)\sim\zeta_0}{e^{- t/2\kappa}\left(\|x_0-y_0\|_2^2 + \|(x_0+v_0)-(y_0+w_0)\|_2^2\right)}\\
%& = e^{- t/2\kappa} W_2( q_0, q^*).
%\end{align*}
%$\zeta_t$ is the coupling from Theorem \ref{t:contraction}.
%The first inequality is due to the fact that $W_2$ uses the optimal coupling, the second inequality is by Theorem \ref{t:contraction}, and the final equality is by definition of $\zeta_0 \in \Gamma_{opt} (q_0,q^*)$. Note the small abuse of notation, since $\zeta_t$ is really a function of $(x_0,v_0,y_0,w_0)$.
\end{Proof}
\begin{lemma}[Sandwich Inequality]\label{l:sandwich}
The triangle inequality for the Euclidean norm implies that
\begin{equation}\label{e:pqsandwich}
\frac{1}{2}W_2(p_t,p^*) \leq W_2(q_t,q^*) \leq 2W_2(p_t,p^*).
\end{equation}
Thus we also get convergence of $\Phi_t p_0$ to $p^*$:
$$W_2(\Phi_t p_0,p^*) \leq 4e^{- t/2\kappa} W_2(p_0,p^*).$$
%The above is inequality is not used anywhere in this paper.
\end{lemma}

\begin{Proof}[Proof of Lemma \ref{l:sandwich}]
Using Young's inequality, we have 
$$\|x+v - (x'+v')\|_2^2 \leq 2\|x-x'\|_2^2 +2\|v-v'\|_2^2.$$
Let $\gamma_t \in \Gamma_{opt} (p_t,p^*).$ Then
\begin{align*}
W_2(q_t,q^*) 
&\leq \sqrt{\Ep{(x,v,x',v')\sim\gamma_t}{\|x-x'\|_2^2 + \|x+v-(x'+v')\|_2^2}}\\
&\leq \sqrt{\Ep{(x,v,x',v')\sim\gamma_t}{3\|x-x'\|_2^2 + 2\|v-v'\|_2^2}}\\
&\leq 2\sqrt{\Ep{(x,v,x',v')\sim\gamma_t}{\|x-x'\|_2^2 + \|v-v'\|_2^2}}\\
&= 2 W_2(p_t,p^*).
\end{align*}
The other direction follows identical arguments, using instead the inequality
$$\|v-v'\|_2^2 \leq 2\|x+v - (x'+v')\|_2^2 + 2\|x-x'\|_2^2.$$
\end{Proof}

We now turn to the proof of Theorem \ref{t:contraction}.

\begin{Proof}[Proof of Theorem \ref{t:contraction}]
We will prove Theorem \ref{t:contraction} in four steps. Our proof relies on a synchronous coupling argument, where $p_t$ and $p_t'$ are coupled (trivially) through independent $p_0$ and $p_0'$, and through shared Brownian motion $B_t$.

\textbf{Step 1:}
Following the definition of \eqref{e:exactlangevindiffusion}, we get
\begin{align*}
\ddt \lrb{(x_t+v_t) - (y_t+w_t)}
=& -(\gamma -1)v_t - u \nabla f(x_t) - \left \{-(\gamma -1)w_t - u \nabla f(y_t)\right \}.
%=& -[(x_t+v_t) -(y_t+u_t)] \\
%&\qquad \qquad + [(x_t-u \nabla f(x_t)) - (y_t - u \nabla f(y_t))].
\end{align*}
The two processes are coupled synchronously which ensures that the Brownian motion terms cancel out. For simplicity, we define $z_t\triangleq x_t - y_t$ and $\psi_t\triangleq v_t - w_t$. As $f$ is twice differentiable, by Taylor's theorem we have 
\begin{align*}
\nabla f(x_t) - \nabla f(y_t) = \underbrace{\left[\int_0^1 \nabla^2 f(x_t+h(y_t-x_t))dh\right]}_{\triangleq\mathcal{H}_t}z_t. 
\end{align*}
Using the definition of $\mathcal{H}_t$ we obtain
\begin{align*}
\ddt \lrb{z_t+\psi_t} =& -( (\gamma -1) \psi_t + u \mathcal{H}_t z_t).
\end{align*}
Similarly we also have the following derivative for the position update:
\begin{align*}
\ddt \lrb{x_t - y_t} & = \ddt \lrb{z_t} = \psi_t.
\end{align*}

\textbf{Step 2:} Using the result from Step 1, we get
\begin{align}
\nonumber &\ddt \left[\eu{z_t+\psi_t}_2^2 + \eu{z_t}_2^2\right]  \\
\nonumber & =  -2\langle(z_t+\psi_t,z_t),((\gamma-1)\psi_t + u\mathcal{H}_t z_t,-\psi_t) \rangle\\
\label{e:contractimp}& = -2 \begin{bmatrix}
  z_t + \psi_t & z_t \\
 \end{bmatrix}\underbrace{\begin{bmatrix}
  (\gamma - 1) I_{d\times d} & u \mathcal{H}_t - (\gamma - 1)I_{d\times d} \\
  -I_{d\times d} & I_{d\times d}
 \end{bmatrix}}_{S_t}\begin{bmatrix}
  z_t + \psi_t \\
  z_t 
 \end{bmatrix}
\end{align}

Here $(z_t + \psi_t, z_t)$ denotes the concatenation of $z_t+\psi_t$ and $z_t$.
% simplicity, we define $z_t\triangleq x_t - y_t$, $w_t\triangleq v_t - u_t$ and $g_t \triangleq -(u \nabla f(x_t) - u \nabla f(y_t))$. Then
%\begin{align*}
%\ddt \eu{(x_t+v_t) - (y_t+u_t)}_2^2
%& = -2 \lin{z_t+w_t, z_t+w_t} + 2\lin{z_t+w_t, z_t+g_t}\\
%& =  - \|z_t+w_t\|_2^2  - \|z_t+w_t\|_2^2 + 2\lin{z_t+w_t,z_t+g_t}\\
%& \le -\|z_t+w_t\|_2^2  + \|z_t+g_t\|_2^2
%\end{align*}
%where in the last line we used $-\|b\|_2^2 + 2\lin{b,c}\leq \|c\|_2^2$ for any two vectors $b$ and $c$. Let us now focus on the $\|z_t+g_t\|_2^2$ term, which is really
%$$\left\|x_t-y_t - u\left( \nabla f(x_t) -  \nabla f(y_t)\right)\right\|_2^2$$
%Lemma \ref{t:graddescentlem} motivates the choice of $u=1/L$ and we get the bound 
%$$\left\|x_t-y_t - \left(\frac{1}{L} \nabla f(x_t) - \frac{1}{L} \nabla f(y_t)\right)\right\|_2^2\leq (1-\alpha)\|x_t-y_t\|_2^2$$ 
%where $\alpha = m/L$ as previously defined. Putting this together we get
%%\begin{align*}
%\ddt \eu{(x_t+v_t) - (y_t+u_t)}_2^2
%\leq& -\|(x_t-y_t) + (v_t-u_t)\|_2^2 + (1-\alpha) \|x_t-y_t\|_2^2\\
%=& -2 \lin{v_t-u_t,x_t-y_t} - \|v_t-u_t\|_2^2- \alpha \|x_t-y_t\|_2^2.
%\end{align*}
%
%

\textbf{Step 3:} Note that for any vector $x\in \mathbb{R}^{2d}$ the quadratic form $x^{\top}S_t x$ is equal to 
\begin{align*}
x^{\top}S_t x & = x^{\top}\left( \frac{S_t + S_t^{\top}}{2}\right)x.
\end{align*}
Let us define the symmetric matrix $Q_t = (S_t + S_t^{\top})/2$. We now compute and lower bound the eigenvalues of the matrix $Q_t$ by making use of an appropriate choice of the parameters $\gamma$ and $u$. The eigenvalues of $Q_t$ are given by the characteristic equation
\begin{align*}
\det\left(\begin{bmatrix}
  (\gamma - 1- \lambda) I_{d\times d} & \frac{u \mathcal{H}_t - (\gamma)I_{d\times d}}{2} \\
  \frac{u \mathcal{H}_t - (\gamma)I_{d\times d}}{2} & (1-\lambda) I_{d\times d}
 \end{bmatrix} \right) & = 0.
\end{align*}
By invoking a standard result of linear algebra (stated in the Appendix as  Lemma \ref{t:blockmatrix}), this is equivalent to solving the equation
\begin{align*}
\det\left((\gamma - 1-\lambda)(1-\lambda)I_{d\times d}- \frac{1}{4}\left( u \mathcal{H}_t - \gamma I_{d\times d}\right)^2\right) = 0.
\end{align*}
Next we diagonalize $\mathcal{H}_t$ and get $d$ equations of the form
\begin{align*}
(\gamma - 1  -\lambda )(1-\lambda )- \frac{1}{4}\left( u \Lambda_j - \gamma\right)^2 = 0,
\end{align*}
where $\Lambda_j$ with $j\in \{1,\ldots d\}$ are the eigenvalues of $\mathcal{H}_t$. By the strong convexity and smoothness assumptions we have $0<m\le\Lambda_j \le L$. We plug in our choice of parameters, $\gamma = 2$ and $u= 1/L$, to get the following solutions to the characteristic equation:
\begin{align*}
\lambda^*_j = 1 \pm \left(1 - \frac{\Lambda_j}{2L}\right).
\end{align*}
This ensures that the minimum eigenvalue of $Q_t$ satisfies $\lambda_{min}(Q_t) \ge 1/2\kappa$. 
%\begin{align*}
%\lambda_j^* & = \frac{\gamma \pm \sqrt{\gamma^2 - 4 u \Lambda_j}}{2}.
%\end{align*}
%Our choice of $u = 1/L$ and $\gamma = 2\sqrt{\kappa}$ ensures that all the eigenvalues of the characteristic equation are complex with real part $\gamma/2 = \sqrt{\kappa}$ and with magnitude equal to $\Lambda_j/L$

\textbf{Step 4:}
Putting this together with our results in Step 2 we have the lower bound
\begin{align*}
\left[z_t+\psi_t,z_t \right]^{\top}S_t \left[z_t+\psi_t,z_t \right] & = \left[z_t+\psi_t,z_t \right]^{\top} Q_t \left[z_t+\psi_t,z_t \right]\\
& \ge \frac{1}{2\kappa}\left[ \lVert z_t+\psi_t  \rVert_2^2 + \lVert z_t \rVert_2^2\right].
\end{align*}
Combining this with \eqref{e:contractimp} yields
\begin{align*}
\ddt \left[\eu{z_t+ \psi_t}_2^2 + \eu{z_t}_2^2\right] & \le - \frac{1}{\kappa}\left[ \lVert z_t+\psi_t  \rVert_2^2 + \lVert z_t \rVert_2^2\right].
\end{align*}
%
%Summing together the two expressions from Step 2 and Step 3, we get
%\begin{align*}
%&\ddt \lrb{\|x_t-y_t\|_2^2 +\eu{(x_t+v_t) - (y_t+u_t)}_2^2}\\
%\leq & -\|v_t-u_t\|_2^2 -\alpha  \|x_t-y_t\|_2^2\\
%\le & -\alpha \left[\lv v_t - u_t \rv_2^2 + \lv x_t - y_t\rv_2^2 \right]\\
%\le & - \frac{\alpha}{2} \left[ \left(\lv v_t - u_t \rv_2^2 + \lv x_t - y_t\rv_2^2\right)+ \left( \lv x_t - y_t\rv_2^2\right)\right]\\
%\leq & -\frac{\alpha}{4} \lrb{\|x_t-y_t\|_2^2 + \eu{(x_t+v_t) - (y_t+u_t)}_2^2}
%\end{align*}
%where in the last step we used Young's inequality that 
%$$\eu{(x_t+v_t) - (y_t+u_t)}_2^2\leq 2\|x_t-y_t\|_2^2 + 2\|v_t-u_t\|_2^2$$
%and the fact that $\alpha\leq 1$ by definition. 
The convergence rate of Theorem \ref{t:contraction} follows immediately from this result by applying Gr{\"o}nwall's inequality (Corollary 3 in \citep{dragomir2000some}).
\end{Proof}
\section{Discretization Analysis}
\label{s:discreteerror}
In this section, we study the solutions of the discrete process \eqref{e:discretelangevindiffusion} up to $t=\delta$ for some small $\delta$. Here, $\delta$ represents a \emph{single step} of the Langevin MCMC algorithm. In Theorem \ref{t:discretizationerror}, we will bound the discretization error between the continuous-time process \eqref{e:exactlangevindiffusion} and the discrete process \eqref{e:discretelangevindiffusion} starting from the same initial distribution.
In particular, we bound $W_2 (\Phi_\delta p_0,\Phit_\delta p_0 )$. This will be sufficient to get the convergence rate stated in Theorem \ref{t:kstepconvergence}. Recall the definition of $\Phi_t$ and $\Phit_t$ from \eqref{d:phi}.

Furthermore, we will assume for now that the kinetic energy (second moment of velocity) is bounded for the continuous-time process,
\begin{equation}
\label{e:energyisbounded}
\forall t\in [0,\d]\quad \Ep{p_t} {\|v\|_2^2}\leq \ke.
\end{equation}
We derive an explicit bound on $\ke$ (in terms of problem parameters $d,L,m$ etc.) in Lemma \ref{l:kineticenergyisbounded} in Appendix \ref{s:kineticenergybound}. 

In this section, we will repeatedly use the following inequality:
$$\eu{\int_0^t v_s ds}_2^2  =  \eu{\frac{1}{t}\int_0^t t \cdot v_s ds}_2^2 \leq t\int_0^t \|v_s\|_2^2 ds,$$
which follows from Jensen's inequality using the convexity of $\|\cdot\|_2^2$.  

We now present our main discretization theorem:
\begin{theorem}\label{t:discretizationerror}
Let $\Phi_t$ and $\Phit_t$ be as defined in \eqref{d:phi} corresponding to the continuous-time and discrete-time processes respectively. Let $p_0$ be any initial distribution and assume wlog that the step size $\delta\le 1$. As before we choose $u = 1/L$ and $\gamma = 2$. Then the distance between the continuous-time process and the discrete-time process is upper bounded by
$$W_2(\Phi_\d p_0,\Phit_\d p_0)\leq \d^2 \sqrt{\frac{2\ke}{5}}.$$
\end{theorem}
\begin{Proof} We will once again use a standard synchronous coupling argument, in which $\Phi_\d p_0$ and $\Phit_\d p_0$ are coupled through the same initial distribution $p_0$ and common Brownian motion $B_t$. 

First, we bound the error in velocity. By using the expression for $v_t$ and $\tilde{v}_t$ from Lemma \ref{l:explicitform}, we have 
\begin{align*}
\E{\eu{v_s -\vt_s }_2^2 } & \overset{(i)}{=} \mathbb{E} \left[ \left\lv u \int_0^s e^{-2(s-r)}\left( \nabla f(x_r) - \nabla f(x_0) \right)dr\right\rv_2^2\right]\\
& = u^2 \mathbb{E} \left[ \left\lv  \int_0^s e^{-2(s-r)}\left( \nabla f(x_r) - \nabla f(x_0) dr\right)\right\rv_2^2\right]\\
& \overset{(ii)}{\le} s u^2 \int_0^s \mathbb{E} \left[ \left\lv e^{-2(s-r)}\left( \nabla f(x_r) - \nabla f(x_0) \right)\right\rv_2^2\right]dr\\
& \overset{(iii)}{\le} s u^2 \int_0^s \mathbb{E} \left[ \left\lv \left( \nabla f(x_r) - \nabla f(x_0) \right)\right\rv_2^2\right]dr\\
& \overset{(iv)}{\le} s u^2 L^2 \int_0^s \mathbb{E} \left[\left \lv x_r - x_0 \right \rv_2^2\right]dr\\
& \overset{(v)}{=} s u^2 L^2 \int_0^s  \mathbb{E} \left[\left \lv \int_0^r v_w dw \right \rv_2^2\right]dr\\
& \overset{(vi)}{\le} s u^2 L^2\int_0^s r \left(\int_0^r \mathbb{E}\left[ \lv v_w \rv_2^2\right] dw\right)dr\\
& \overset{(vii)}{\le} s u^2 L^2 \ke \int_0^s r \left(\int_0^r dw\right) dr\\
& = \frac{s^4 u^2 L^2 \ke}{3},
\end{align*}
where $(i)$ follows from the Lemma \ref{l:explicitform} and $v_0=\vt_0$, $(ii)$ follows from application of Jensen's inequality, $(iii)$ follows as $\lvert e^{-4(s-r)}\rvert \le 1$, $(iv)$ is by application of the $L$-smoothness property of $f(x)$, $(v)$ follows from the definition of $x_r$, $(vi)$ follows from Jensen's inequality and $(vii)$ follows by the uniform upper bound on the kinetic energy assumed in \eqref{e:energyisbounded}, and proven in Lemma \ref{l:kineticenergyisbounded}.

This completes the bound for the velocity variable. Next we bound the discretization error in the position variable:
%\begin{align*}
%& \E{\eu{v_s -\vt_s }_2^2 }\\
%=& \E{\eu{\int_0^s (-2 v_r - \frac{1}{L}\nabla f(x_r)) - (-2 \vt_r - \frac{1}{L} \nabla f(\xt_0))dr }_2^2}\\
%\leq& 2\E{\eu{\int_0^s v_r - \vt_r dr}_2^2} + 2 \E{\eu{\int_0^s\frac{1}{L} \nabla f(x_r) - \frac{1}{L} \nabla f(\xt_r) dr}_2^2}\\
%\leq& 2s\int_0^s\E{\eu{ v_r - \vt_r }_2^2}dr + 2s\int_0^s \E{\eu{\frac{1}{L} \nabla f(x_r) - \frac{1}{L} \nabla f(\xt_r) }_2^2} dr\\
%\leq& 2s^2\E{\eu{ v_s - \vt_s }_2^2} + 2s\int_0^s \E{\eu{\frac{1}{L} \nabla f(x_r) - \frac{1}{L} \nabla f(\xt_r) }_2^2} dr
%\end{align*}
%Rearranging the terms, and assuming that $s\leq \delta \leq \frac{1}{2}$, we get
%\begin{align*}
%& \E{\eu{v_s -\vt_s }_2^2 }\\
%\leq& \frac{4s}{L^2}\int_0^s \E{\eu{ \nabla f(x_r) - \nabla f(\xt_r) }_2^2} dr\\
%\leq&  4s \int_0^s \E{\eu{x_r-x_0}_2^2} dr\\
%=& 4s \int_0^s \E{\eu{\int_0^r v_t dt}_2^2} dr\\
%\leq& 4s \int_0^s r \int_0^r \E{\eu{v_t}_2^2} dt dr\\
%\leq & 4 s^4 C
%\end{align*}
%
%Where the third line is by $L$-lipschitz continuous gradients, the fourth line is by \eqref{e:energyisbounded}, the fifth line is by Jensens, and the last line is by the assumption in \eqref{e:energyisbounded}.
\begin{align*}
\E{\eu{x_s - \xt_s}_2^2} & = \E{\eu{\int_0^s (v_r - \vt_r) dr}_2^2}\\
& \leq s\int_0^s \mathbb{E} \left[\lv v_r - \tilde{v}_r \rv_2^2\right]dr\\
& \le s \int_0^s \frac{r^4u^2L^2\ke}{3}dr\\
& = \frac{s^6 u^2 L^2 \ke }{15},
\end{align*}
where the first line is by coupling through the initial distribution $p_0$, the second line is by Jensen's inequality and the third inequality uses the preceding bound. Setting $s = \delta$ and by our choice of $u= 1/L$ we have that the squared Wasserstein distance is bounded as
\begin{align*}
W^2_2(\Phi_\delta p_0,\tilde{\Phi} p_0) \le \ke \left(\frac{\delta^4}{3}+ \frac{\delta^6}{15}\right).
\end{align*}
Given our assumption that $\delta$ is chosen to be smaller than 1, this gives the upper bound:
\begin{align*}
W^2_2(\Phi_\delta p_0,\tilde{\Phi} p_0) \le \frac{2\ke \delta^4}{5}.
\end{align*}
Taking square roots establishes the desired result.
\end{Proof}
\section{Proof of Theorem \ref{t:kstepconvergence}}
\label{s:discreteconvergence}
Having established the convergence rate for the continuous-time SDE \eqref{e:exactlangevindiffusion} and having proved a discretization error bound in Section \ref{s:discreteerror} we now put these together and establish our main result for underdamped Langevin MCMC.%We will now prove the convergence rate of the second order Langevin MCMC by studying the \textbf{single step} convergence rate of the discretized SDE \eqref{e:discretelangevindiffusion}. 
\begin{Proof}[Proof of Theorem \ref{t:kstepconvergence}]
From Corollary \ref{c:exactconvergence}, we have that for any $i \in \{1,\ldots,n \}$
%$$W_2(\Phi_\d q^{(i)},q^*)\leq e^{-\kappa \delta/2}W_2(q^{(i)}, q^*) \le \left(1- \frac{\kappa\delta}{4}\right)W_2(q^{(i)}, q^*)$$
$$W_2(\Phi_\d q^{(i)},q^*)\leq e^{- \delta/2\kappa}W_2(q^{(i)}, q^*).$$
%the second inequality follows as $\kappa\delta/2 < 1$. 
By the discretization error bound in Theorem \ref{t:discretizationerror} and the sandwich inequality \eqref{e:pqsandwich}, we get
$$W_2(\Phi_\d q^{(i)}, \Phit_\d q^{(i)})\leq 2W_2(\Phi_\d p^{(i)}, \Phit_\d p^{(i)})\leq \d^2 \sqrt{\frac{8\ke}{5}}.$$
By the triangle inequality for $W_2$, 
%\begin{align}\label{e:singlestepdiscreteimprovement}
%W_2(q^{(i+1)}, q^*) = W_2(\Phit_\d q^{(i)}, q^*) & \le W_2(\Phi_\d q^{(i)}, \Phit_\d q^{(i)}) + W_2(\Phi_\d q^{(i)},q^*)\\
%& \le \d^2 \sqrt{\frac{8\ke}{5}} + \left(1- \frac{\kappa\delta}{4}\right)W_2(q^{(i)}, q^*).
%\end{align}
\begin{align}\label{e:singlestepdiscreteimprovement}
W_2(q^{(i+1)}, q^*) = W_2(\Phit_\d q^{(i)}, q^*) & \le W_2(\Phi_\d q^{(i)}, \Phit_\d q^{(i)}) + W_2(\Phi_\d q^{(i)},q^*)\\
& \le \d^2 \sqrt{\frac{8\ke}{5}} + e^{-\delta/2\kappa}W_2(q^{(i)}, q^*) \label{e:singlestepdiscreteimprovement}.
\end{align}
Let us define $\eta = e^{-\delta/ 2\kappa}$. Then by applying \eqref{e:singlestepdiscreteimprovement}  $n$ times we have:
\begin{align*}
W_2(q^{(n)}, q^*) & \le \eta^{n}W_2(q^{(0)}, q^*)+ \left(1+ \eta + \ldots
+ \eta^{n-1} \right)\d^2 \sqrt{\frac{8\ke}{5}}\\ 
& \le 2\eta^{n}W_2(p^{(0)}, p^*)+ \left(\frac{1}{1-\eta}\right)\d^2 \sqrt{\frac{8\ke}{5}},
\end{align*}
where the second step follows by summing the geometric series and by applying the upper bound \eqref{e:pqsandwich}. By another application of \eqref{e:pqsandwich} we get:
\begin{align*}
W_2(p^{(n)}, p^*) & \le \underbrace{4\eta^{n}W_2(p^{(0)}, p^*)}_{T_1}+ \underbrace{\left(\frac{1}{1-\eta}\right)\d^2 \sqrt{\frac{32\ke}{5}}}_{T_2}.
\end{align*}
Observe that 
\begin{align*}
1-\eta = 1-e^{-\delta/2\kappa} & \ge \frac{\delta}{4\kappa}.
\end{align*}
This inequality follows as $ \delta/\kappa <1$. We now bound both terms $T_1$ and $T_2$ at a level $\varepsilon/2$ to bound the total error $W_2(p^{(n)},p^*)$ at a level $\varepsilon$. Note that choice of $\delta = \varepsilon \kappa^{-1} \sqrt{1/10816\left(d/m + \mathcal{D}^2 \right)} \le \varepsilon \kappa^{-1}  \sqrt{5 /2048 \ke}$ (by upper bound on $\ke$ in Lemma \ref{l:kineticenergyisbounded}) ensures that,
\begin{align*}
T_2 = \left(\frac{1}{1-\eta}\right)\d^2 \sqrt{\frac{32\ke}{5}} \le \frac{4\kappa}{ \delta}\left(\delta^2 \sqrt{\frac{32\ke}{5}} \right) \le \frac{\varepsilon}{2}.
\end{align*}
To control $T_1< \varepsilon/2$ it is enough to ensure that
\begin{align*}
n > \frac{1}{\log(\eta)} \log \left(\frac{8 W_2(p^{(0)},p^*)}{\varepsilon} \right).
\end{align*}
In Lemma \ref{lem:initialdistancebound} we establish a bound on $W_2^2(p^{(0)},p^*) \le 3(d/m + \mathcal{D}^2)$. This motivates our choice of $n > \frac{\kappa}{ \delta}\log\left( \frac{24 \left(\frac{d}{m}+ \mathcal{D}^2 \right)}{\varepsilon} \right)$, which establishes our claim.
%Next if we choose $n > \frac{1}{\kappa \delta}\log(8W_2(p^{(0)},p^*)/\varepsilon)$ ensures that $T_1$ is bounded
%As long as $W_2(q^{(i)}, q^*)\geq \varepsilon/2$ and $\delta = \varepsilon \kappa \left( \sqrt{\frac{5 }{2048 \ke}}\right)$, we get a one-step-contraction for Algorithm \ref{algosec}
%\begin{align*}
%W_2(q^{(i+1)}, q^*) \le \left(1- \frac{\kappa \delta}{8}\right)W_2(q^{(i)}, q^*).
%\end{align*}
%Repeatedly applying the above inequality for $n$ steps, we get
%\begin{align*}
%W_2(q^{(n)}, q^*) \le \left(1- \frac{\kappa \delta}{8}\right)^{n}W_2(q^{(0)}, q^*) \le \exp\left(-\frac{n\kappa\delta}{8}\right)W_2(q^{(0)}, q^*) .
%\end{align*}
%To satisfy
%\begin{align*}
%W_2(q^{(n)} , q^*)\leq \frac{\varepsilon}{2}
%\end{align*}
%we choose $n > \frac{8}{\kappa \delta} \log\left( \frac{4 W_2(p^{(0)},p^*)}{\varepsilon}\right)$. The conclusion
%$$W_2(p^{(n)} , p^*)\leq \varepsilon$$
%follows from \eqref{e:pqsandwich}, completing the proof.
\end{Proof}
\section{Conclusion}
We present an MCMC algorithm based on the underdamped Langevin diffusion and provide guarantees for its convergence to the invariant distribution in 2-Wasserstein distance. Our result is a quadratic improvement in both dimension ($\sqrt{d}$ instead of $d$) as well as error ($1/\varepsilon$ instead of $1/\varepsilon^2$) for sampling from strongly log-concave distributions compared to the best known results for overdamped Langevin MCMC. In its use of underdamped, second-order dynamics, our work also has connections to Nesterov acceleration \citep{nesterov} and to Polyak's heavy ball method \citep{polyak}, and adds to the growing body of work that aims to understand acceleration of first-order methods as a discretization of continuous-time processes. 

An interesting open question is whether we can improve the dependence on the condition number from $\kappa^2$ to $\kappa$. Another interesting direction would to explore if our approach can be used to sample efficiently from \emph{non-log-concave} distributions. Also, lower bounds in the MCMC field are largely unknown and it would extremely useful to understand the gap between existing algorithms and optimal achievable rates. Another question could be to explore the wider class of second-order Langevin equations and study if their discretizations provide better rates for sampling from particular distributions.
%\subsubsection*{Acknowledgments}
%\input{ack}
\nocite{*}
\bibliography{ref} 
\appendix
\newpage
\section{Explicit Discrete Time Updates}
\label{s:howtodiscretize}
In this section we calculate integral representations of the solutions to the continuous-time process \eqref{e:exactlangevindiffusion} and the discrete-time process \eqref{e:discretelangevindiffusion}.
\begin{lemma}\label{l:explicitform}
The solution $(x_t,v_t)$ to the underdamped Langevin diffusion \eqref{e:exactlangevindiffusion} is
\begin{align*}
\numberthis \label{e:vdynamics}
v_t &= v_0 e^{-\gamma t} - u \left(\int_0^t e^{-\gamma(t-s)} \nabla f(x_s) ds \right) + \sqrt{2\gamma u} \int_0^t e^{-\gamma (t-s)} dB_s\\
x_t &= x_0 + \int_0^t v_s ds.
\end{align*}
The solution $(\tilde{x}_t,\tilde{v}_t)$ of the discrete underdamped Langevin diffusion \eqref{e:discretelangevindiffusion} is
\begin{align*}
\numberthis \label{e:vtildedynamics}
\vt_t &= \vt_0 e^{-\gamma t} - u \left(\int_0^t e^{-\gamma(t-s)} \nabla f(\xt_0) ds \right) + \sqrt{2\gamma u} \int_0^t e^{-\gamma (t-s)} dB_s\\
\xt_t &= \xt_0 + \int_0^t \vt_s ds.
\end{align*}
\end{lemma}
\begin{Proof}%[Proof of Lemma \ref{l:explicitform}]
It can be easily verified that the above expressions have the correct initial values $(x_0,v_0)$ and $(\xt_0,\vt_0)$. By taking derivatives, one also verifies that they satisfy the differential equations in \eqref{e:exactlangevindiffusion} and \eqref{e:discretelangevindiffusion}.
\end{Proof}
Next we calculate the moments of the Gaussian used in the updates of Algorithm \ref{ulmcmc}. These are obtained by integrating the expression for the discrete-time process presented in Lemma \ref{l:explicitform}.
\begin{lemma}\label{l:gaussianexpressionforsamplingxtvt}
Conditioned on $(\tilde{x}_0,\tilde{v}_0)$, the solution $(\xt_t,\vt_t)$ of \eqref{e:discretelangevindiffusion} with $\gamma=2$ and $u=1/L$ is a Gaussian with conditional mean,
\begin{align*}
\E{\vt_t } & = \vt_0 e^{-2 t} - \frac{1}{2L}(1-e^{-2 t}) \nabla f(\xt_0)\\
\E{\xt_t } & = \xt_0 + \frac{1}{2}(1-e^{-2 t})\vt_0 - \frac{1}{2L} \left( t - \frac{1}{2}\left(1-e^{-2 t}\right) \right) \nabla f(\xt_0),
\end{align*}
and with conditional covariance,
\begin{align*}
\E{\left(\xt_t - \E{\xt_t}\right) \left(\xt_t - \E{\xt_t}\right)^{\top}  }&= \frac{1}{L } \left[t-\frac{1}{4}e^{-4t}-\frac{3}{4}+e^{-2t}\right] \cdot I_{d\times d}\\
\E{\left(\vt_t - \E{\vt_t}\right) \left(\vt_t - \E{\vt_t}\right)^{\top} } &= \frac{1}{L}(1-e^{-4 t})\cdot I_{d\times d}\\
\E{\left(\xt_t - \E{\xt_t}\right) \left(\vt_t - \E{\vt_t}\right)^{\top} }&= \frac{1}{2L} \left[1+e^{-4t}-2e^{-2t}\right] \cdot I_{d \times d}.
\end{align*}
\end{lemma}
\begin{Proof}%[Proof of Lemma \ref{l:gaussianexpressionforsamplingxtvt}]
It follows from the definition of Brownian motion that the distribution of $(\tilde{x}_t,\tilde{v}_t)$ is a $2d$-dimensional Gaussian distribution. We will compute its moments below, using the expression in Lemma \ref{l:explicitform} with $\gamma=2$ and $u=1/L$.

Computation of the conditional means is straightforward, as we can simply ignore the zero-mean Brownian motion terms:
\begin{align}
\E{\vt_t } & = \vt_0 e^{-2 t} - \frac{1}{2L}(1-e^{-2 t}) \nabla f(\xt_0)\\
\E{\xt_t } & = \xt_0 + \frac{1}{2}(1-e^{-2 t})\vt_0 - \frac{1}{2L} \left( t - \frac{1}{2}\left(1-e^{-2 t}\right) \right) \nabla f(\xt_0).
\end{align}

The conditional variance for $\vt_t$ only involves the Brownian motion term:
\begin{align*}
&\E{\left(\vt_t - \E{\vt_t}\right) \left(\vt_t - \E{\vt_t}\right)^{\top} }\\
=& \frac{4}{L}\E{\left(\int_0^t e^{-2 (t-s)} dB_s\right)\left(\int_0^t e^{-2(s-t)} dB_s\right)^{\top}}\\
=& \frac{4}{L} \left(\int_0^t e^{-4(t-s)} ds \right) \cdot I_{d\times d}\\
=& \frac{1}{L}(1-e^{-4 t})\cdot I_{d\times d}.
\end{align*}
The Brownian motion term for $\xt_t$ is given by
\begin{align*}
\sqrt{\frac{4}{L}} \int_0^t  \left( \int_0^r e^{-2 (r-s)} dB_s \right)dr
=& \sqrt{\frac{4}{L}} \int_0^t e^{2s}\left( \int_s^t e^{-2r} dr \right) dB_s\\
=& \sqrt{\frac{1}{L}} \int_0^t \left(1-e^{-2(t-s)} \right) dB_s.
\end{align*}
Here the second equality follows by Fubini's theorem. The conditional covariance  for $\tilde{x}_t$ now follows as 
\begin{align*}
&\E{\left(\xt_t - \E{\xt_t}\right) \left(\xt_t - \E{\xt_t}\right)^{\top} }\\
=& \frac{1}{L } \E{\left( \int_0^t \left(1- e^{-2(t-s)}  \right)dB_s\right)\left( \int_0^t \left(1-e^{-2(t-s)} \right)dB_s\right)^{\top}}\\
=& \frac{1}{L } \left[ \int_0^t \left(1-e^{-2(t-s)}\right )^2 ds \right]\cdot I_{d \times d}\\
=& \frac{1}{L } \left[t-\frac{1}{4}e^{-4t}-\frac{3}{4}+e^{-2t}\right] \cdot I_{d\times d}.
\end{align*}
Finally we compute the cross-covariance between $\tilde{x}_t$ and $\tilde{v}_t$,
\begin{align*}
&\E{\left(\xt_t - \E{\xt_t}\right) \left(\vt_t - \E{\vt_t}\right)^{\top}}\\
=& \frac{2}{L} \E{\left( \int_0^t\left(1-e^{-2 (t-s)} \right)dB_s\right)\left( \int_0^t e^{-2 (t-s)} dB_s\right)^{\top}}\\
=& \frac{2}{L} \left[ \int_0^t(1-e^{-2 (t-s)})(e^{-2 (t-s)}) ds \right]\cdot I_{d \times d}\\
=& \frac{1}{2L} \left[1+e^{-4t}-2e^{-2t}\right] \cdot I_{d \times d}.
\end{align*}

We thus have an explicitly defined Gaussian. Notice that we can sample from this distribution in time linear in $d$, since all $d$ coordinates are independent.
\end{Proof}
\section{Controlling the Kinetic Energy}
\label{s:kineticenergybound}
In this section, we establish an explicit bound on the kinetic energy $\ke$ in \eqref{e:energyisbounded} which is used to control the discretization error at each step.

\begin{lemma}[Kinetic Energy Bound]\label{l:kineticenergyisbounded}
Let $p^{(0)}(x,v) =1_{x=x^{(0)}} \cdot 1_{v=0}$--- the Dirac delta distribution at $(x^{(0)},0)$. Let the initial distance from the optimum satisfy $\lv x^{(0)}-x^* \rv_2^2 \le \mathcal{D}^2$ and $u=1/L$ as before. Further let $p^{(i)}$ be defined as in Theorem \ref{t:kstepconvergence} for $i=1,\ldots n$, with step size $\d$ and number of iterations $n$ as stated in Theorem \ref{t:kstepconvergence}. Then for all $i=1,\ldots n$ and for all $t\in[0,\d]$, we have the bound
$$\Ep{(x,v)\sim \Phi_t p^{(i)}}{\|v\|_2^2}\leq \ke,$$
with $\ke = 26(d/m +  \mathcal{D}^2)$.
\end{lemma}
\begin{Proof} We first establish an inequality that provides an upper bound on the kinetic energy for any distribution $p$.

\textbf{Step 1}:  Let $p$ be any distribution over $(x,v)$, and let $q$ be the corresponding distribution over $(x,x+v)$. Let $(x',v')$ be random variables with distribution $p^*$. Further let $\zeta \in \Gamma_{opt}(p,p^*)$ such that,
 $$\mathbb{E}_{\zeta} \left[\lv x-x'\rv_2^2 + \lv (x-x')+(v-v')\rv_2^2 \right] = W_2^2(q,q^*).$$ Then we have,
\begin{align}
\nonumber\Ep{p}{\|v\|_2^2} & = \mathbb{E}_{\zeta}\left[\lv v- v' + v'\rv^2_2\right]\\
\nonumber & \le 2\Ep{p^*}{\|v \|_2^2} + 2  \mathbb{E}_{\zeta}\left[ \lv v- v' \rv^2_2\right]\\
\nonumber & \le 2\Ep{p^*}{\|v \|_2^2} + 4 \mathbb{E}_{\zeta}\left[ \lv x+v - (x'+v') \rv^2_2 + \lv x - x' \rv_2^2 \right]\\
& = 2\Ep{p^*}{\|v \|_2^2} + 4 W_2^2 (q,q^*), \label{e:kineticlessthanwasserstein}
\end{align}
where for the second and the third inequality we have used Young's inequality, while the final line follows by optimality of $\zeta$. 

\textbf{Step 2}: We know that $p^*\propto \exp(-(f(x) + \frac{L}{2}\|v\|_2^2))$, so we have $\Ep{p^*}{\|v\|_2^2} = d/L$. 

\textbf{Step 3}: For our initial distribution $p^{(0)} (q^{(0)})$ we have the bound
\begin{align*}
W_{2}^{2}(q^{(0)},q^*) & \le 2 \mathbb{E}_{p^*} \left[ \lv v \rv_2^2\right] + 2 \mathbb{E}_{x \sim p^{(0)}, x' \sim p^*} \left[\lv x-x' \rv_2^2 \right]\\
& = \frac{2d}{L} + 2 \mathbb{E}_{p^*} \left[ \lv x - x^{(0)} \rv_2^2\right],
\end{align*}
where the first inequality is an application of Young's inequality. The second term is bounded below,
\begin{align*}
\mathbb{E}_{p^*} \left[ \lv x - x^{(0)} \rv_2^2\right] & \le 2\mathbb{E}_{p^*} \left[ \lv x - x^* \rv_2^2\right]+ 2\lv x^{(0)} - x^* \rv_2^2\\
%& = \frac{4}{m}\left( \frac{m}{2}\mathbb{E}_{p^*} \left[ \lv x - x^* \rv_2^2\right]\right)+ 2\lv x^{(0)} - x^* \rv_2^2 \\
%& \le \frac{4}{m} \left( \mathbb{E}_{p^*} \left[ f(x) - f(x^*)\right]\right) + 2\mathcal{D}^2\\
%& \le \frac{4}{m}\left( \frac{d}{2\beta}\right) + 2\mathcal{D}^2 
&\le \frac{2d}{m } + 2\mathcal{D}^2,
\end{align*}
where the first inequality is again by Young's inequality. The second line follows by applying Theorem \ref{t:xvariance} to control $\mathbb{E}_{p^*}\left[\lv x-x^* \rv_2^2\right]$. Combining these we have the bound,
\begin{align*}
W_{2}^{2}(q^{(0)},q^*) \le 2d\left(\frac{1}{L}+ \frac{2}{m} \right) + 4\mathcal{D}^2.
\end{align*}
Putting all this together along with \eqref{e:kineticlessthanwasserstein} we have 
\begin{align*}
\mathbb{E}_{p^{(0)}}\left[ \lv v \rv_2^2\right] & \le \frac{10d}{L} + \frac{16 d}{m} + 16 \mathcal{D}^2\\
& \le 26\left(\frac{ d}{m}+  \mathcal{D}^2\right).
\end{align*}
%To analyze the second term we will first derive a lower bound on the normalizing factor,
%\begin{align*}
%e^{\beta f(x^*)} \int e^{-\beta f(x)} dx & = \int e^{-\beta(f(x) - f(x^*))}dx \\
%& \ge \int e^{-\frac{\beta L }{2}(\lv x-x^* \rv_2^2)} dx = \sqrt{\frac{2\pi}{\beta L}}
%\end{align*}
%where the inequality follows from the L-Lipshitz gradients of $f$. Given this bound we have
%\begin{align*}
%\mathbb{E}_{p^*} \left[ \lv x - x^* \rv_2^2\right] & = \frac{1}{\int e^{-\beta f(x) dx}} \left(\int e^{-\beta f(x)} \lv x- x^* \rv_2^2\right)\\
%& = \frac{1}{\int e^{-\beta (f(x)-f(x^*)) dx}}\left(\int e^{-\beta (f(x)-f(x^*))} \lv x- x^* \rv_2^2\right)\\
%& \le \left(\sqrt{\frac{\beta L}{2 \pi}} \right) \left(\int e^{-\frac{\beta m}{2} \lv x-x^* \rv_2^2} \lv x- x^* \rv_2^2\right) \\
%& \le  \left(\sqrt{\frac{\beta L}{2 \pi}} \right) \left( \frac{d}{\beta m} \sqrt{\frac{2 \pi}{\beta m}} \right)\\
%& = d \sqrt{\frac{L}{\beta^2 m^3}}
%\end{align*}

\textbf{Step 4}:
By Theorem \ref{t:contraction}, we know that $\forall t > 0$, 
\begin{align*}
W_2^2 (\Phi_t q^{(i)}, q^*)\leq W_2^2 (q^{(i)}, q^*). 
\end{align*}
This proves the theorem statement for $i = 0$. We will now prove it for $i > 0$ via induction. We have proved it for the base case $i=0$, let us assume that the result holds for $i>0$. Then by Theorem \ref{t:discretizationerror} applied for $i$ steps, we know that 
$$W_2^2 (q^{(i+1)},q^*) = W_2^2 (\Phit_\d q^{(i)},q^*)\leq W_2^2(q^{(i)},q^*). $$
Thus by \eqref{e:kineticlessthanwasserstein} we have, 
$$\Ep{\Phi_{t} p^{(i)}}{\|v\|_2^2}\leq \ke,$$
for all $t>0$ and $i \in \{0,1,\ldots,n \}$.
\end{Proof}
Next we prove that the distance of the initial distribution $p^{(0)}$ to the optimum distribution $p^*$ is bounded.
\begin{lemma}\label{lem:initialdistancebound} Let $p^{(0)}(x,v) =1_{x=x^{(0)}} \cdot 1_{v=0}$--- the Dirac delta distribution at $(x^{(0)},0)$. Let the initial distance from the optimum satisfy $\lv x^{(0)}-x^* \rv_2^2 \le \mathcal{D}^2$ and $u=1/L$ as before. Then
\begin{align*}
W_2^2(p^{(0)},p^*) \le 3\left( \mathcal{D}^2 + \frac{d}{m} \right).
\end{align*}
\end{lemma}
\begin{Proof} As $p^{(0)}(x,v)$ is a delta distribution, there is only one valid coupling between $p^{(0)}$ and $p^*$. Thus we have
\begin{align*}
W_2^2(p^{(0)},p^*) & = \mathbb{E}_{(x,v) \sim p^*}\left[\lv x-x^{(0)}\rv_2^2 + \lv v \rv_2^2 \right] \\ 
& =\mathbb{E}_{(x,v) \sim p^*}\left[\lv x-x^* + x^* - x^{(0)}\rv_2^2 + \lv v \rv_2^2 \right] \\
& \le 2\mathbb{E}_{x \sim p^*(x)}\left[\lv x-x^*\rv_2^2 \right] + 2\mathcal{D}^2 + \mathbb{E}_{v \sim p^*(v)}\left[\lv v \rv_2^2 \right]
\end{align*}
where the final inequality follows by Young's inequality and by the definition of $\mathcal{D}^2$. Note that $p^*(v) \propto \exp(-L\lv v\rv_2^2/2)$, therefore $\mathbb{E}_{v \sim p^*(v)}\left[\lv v \rv_2^2 \right] = d/L$. By invoking Theorem \ref{t:xvariance} the first term $\mathbb{E}_{x \sim p^*(x)}\left[\lv x-x^*\rv_2^2 \right]$ is bounded by $d/m$. Putting this together we have,
\begin{align*}
W_2^2(p^{(0)},p^*) & \le 2\frac{d}{m} + \frac{d}{L} + 2\mathcal{D}^2 \le 3\left(\frac{d}{m} + \mathcal{D}^2 \right).
\end{align*} 
\end{Proof}
\section{Varying Step Size}
\label{s:stepsizechange}
Here we provide a sharper analysis of underdamped Langevin MCMC by using a varying step size. By choosing an adaptive step size we are able to shave off the log factor appearing in Theorem \ref{t:kstepconvergence}.

\begin{theorem}\label{t:loggone}
Let the initial distribution $p^{(0)}(x,v) =1_{x=x^{(0)}} \cdot 1_{v=0}$ and let the initial distance to optimum satisfy $ \lv x^{(0)} - x^* \rv_2^2 \le \mathcal{D}^2$. Also let $W_2(p^{(0)},p^*) \le 3\left(\frac{d}{m}+\mathcal{D}^2\right) < \epsilon_0$.  We set the initial step size to be  
$$\d_1 = \frac{\epsilon_0 }{2\cdot 104 \kappa} \sqrt{\frac{1}{d/m + \mathcal{D}^2}}, $$
and initial number of iterations,
$$n_1 =  \frac{208\kappa^2}{ \epsilon_0}\cdot\left(\sqrt{\frac{d}{m}+ \mathcal{D}^2}\right)\cdot \log(16).$$
We define a sequence of $\ell$ epochs with step sizes $(\delta_1,\ldots,\delta_{\ell})$ and number of iterations $(n_1,\ldots,n_{\ell})$ where $\delta_1$ and $n_1$ are defined as above. Choose $\ell = \lceil \log(\epsilon^0/\varepsilon)/\log(2)\rceil$ and, for $i\ge 1$ set $\delta_{i+1} = \delta_{i}/2$ and $n_{i+1} = 2 n_{i}$. 

We run $\ell$ epochs of underdamped Langevin MCMC (Algorithm \ref{ulmcmc}) with step size sequence $(\delta_1,\delta_2,\ldots,\delta_{\ell})$ with number of iterations $(n_1,n_2,\ldots,n_{\ell})$ corresponding to each step size. Then we have the guarantee 
\begin{align*}
W_2(p^{(n)} , p^*)\leq \varepsilon,
\end{align*}
with total number of steps $n=n_1+n_2+\ldots+n_{\ell}$ being 
\begin{align*}
n = \frac{416 \log(16)\kappa^2}{ \varepsilon}\cdot \left( \sqrt{\frac{d}{m}+ \mathcal{D}^2}\right).
\end{align*}
\end{theorem}
\begin{Proof} Let the initial error in the probability distribution be $W_2(p^{(0)},p^*) = \epsilon_0$. Then by the results of Theorem \ref{t:kstepconvergence} if we choose the step size to be
\begin{align*}
\delta_1 = \frac{\epsilon_0 }{2\cdot 104 \kappa} \sqrt{\frac{1}{d/m + \mathcal{D}^2}},
\end{align*}
then we have the guarantee that in 
\begin{align*}
n_1 = \frac{208 \kappa^2}{ \epsilon_0}\cdot\left(\sqrt{\frac{d}{m}+ \mathcal{D}^2}\right)\cdot \log(16)
\end{align*}
steps the error will be less than $\epsilon_1 = \epsilon_0/2$. At this point we half the step size $\delta_2 = \delta_1/2$ and run for $n_2 = 2 n_1$ steps. After that we set $\delta_3 = \delta_2/2$ and run for double the steps $n_3= 2 n_2$ and so on. We repeat this for $\ell$ steps. Then at the end if the probability distribution is $p^{(n)}$ by Theorem \ref{t:kstepconvergence} we have the guarantee that $W_2(p^{(n)},p^*) \le \epsilon_0/2^{\ell}<\varepsilon$. The total number of steps taken is 
\begin{align*}
n_1+ n_2 \ldots + n_{\ell}& = \sum_{i=1}^{\ell} n_{i}\\
& = \frac{208\kappa^2}{ \epsilon_0}\cdot \left( \sqrt{\frac{d}{m}+ \mathcal{D}^2}\right)\cdot \log(16) \left\{ \sum_{i=0}^{\ell-1} 2^i\right\}\\
& = 104 \log(16)\kappa^2 \cdot\frac{2^{\ell}}{\epsilon_0}\cdot \left( \sqrt{\frac{d}{m}+ \mathcal{D}^2}\right)\left\{ \sum_{i=0}^{\ell-1} 2^{-i}\right\}\\
& \le 104 \log(16)\kappa^2\cdot\frac{2}{\varepsilon}\cdot \left( \sqrt{\frac{d}{m}+ \mathcal{D}^2}\right)\left\{ 2\right\}\\
& = \frac{416 \log(16)\kappa^2}{\varepsilon}\cdot \left( \sqrt{\frac{d}{m}+ \mathcal{D}^2}\right),
\end{align*}
where the inequality follows by the choice of $\ell$ and an upper bound on the sum of the geometric series.
\end{Proof} 
\section{Analysis with Stochastic Gradients}
\label{app:stochastic}
Here we state the underdamped Langevin MCMC algorithm with stochastic gradients. We will borrow notation and work under the assumptions stated in Section \ref{ss:stochasticgradient}.

\begin{algorithm}[t] \label{alg:ulmcmcnoise}
\caption{Stochastic Gradient Underdamped Langevin MCMC} 
\SetKwInOut{Input}{Input}
    \SetKwInOut{Output}{Output}
        \Input{Step size $\delta<1$, number of iterations $n$, initial point $(x^{(0)},0)$, smoothness parameter $L$ and stochastic gradient oracle $\nh f(\cdot)$}

    \For {$i=0,1,\ldots,n-1$} 
      {
		  Sample $(x^{i+1},v^{i+1})\sim Z^{i+1}(x^i,v^i)$
        }
   \end{algorithm} 
\subsubsection*{Description of Algorithm \ref{alg:ulmcmcnoise}}
The random vector $Z^{i+1}(x_i,v_i)\in \R^{2d}$, conditioned on $(x^i,v^i)$, has a Gaussian distribution with conditional mean and covariance obtained from the following computations:
   \begin{align*}
&\E{v^{i+1}} = v^i e^{-2 \d} - \frac{1}{2L}(1-e^{-2 \d}) \nh f(x^i)\\
&\E{x^{i+1}}  = x^i + \frac{1}{2}(1-e^{-2 \d})v^i - \frac{1}{2L} \left( \d - \frac{1}{2}\left(1-e^{-2 \d}\right) \right) \nh  f(x^i)\\
&\E{\left(x^{i+1} - \E{x^{i+1}}\right) \left(x^{i+1} - \E{x^{i+1}}\right)^{\top}}= \frac{1}{L } \left[\d-\frac{1}{4}e^{-4\d}-\frac{3}{4}+e^{-2\d}\right] \cdot I_{d\times d}\\
&\E{\left(v^{i+1} - \E{v^{i+1}}\right) \left(v^{i+1} - \E{v^{i+1}}\right)^{\top}} = \frac{1}{L}(1-e^{-4 \d})\cdot I_{d\times d}\\
&\E{\left(x^{i+1} - \E{x^{i+1}}\right) \left(v^{i+1} - \E{v^{i+1}}\right)^{\top}}= \frac{1}{2L} \left[1+e^{-4\d}-2e^{-2\d}\right] \cdot I_{d \times d}.
\end{align*}   
 The distribution is obtained by integrating the discrete underdamped Langevin diffusion \eqref{e:discretelangevindiffusionwithnoise} up to time $\delta$, with the specific choice of $\gamma=2$ and $u=1/L$. In other words, if $p^{(i)}$ is the distribution of $(x^i,v^i)$, then $Z^{i+1}(x^i,v^i) \sim p^{(i+1)} = \Phih_{\d} p^{(i)}$. Derivation is identical to the calculation in Appendix \ref{s:howtodiscretize} by replacing exact gradients $\nabla f(\cdot)$ with stochastic gradients $\nh f(\cdot)$. A key ingredient as before in understanding these updates is the next lemma which calculates the exactly the update at each step when we are given stochastic gradients.
\begin{lemma}\label{lem:exactupdatesstochasticgradients}
The solution $(\hat{x}_t,\hat{v}_t)$ of the stochastic gradient underdamped Langevin diffusion \eqref{e:discretelangevindiffusionwithnoise} is
\begin{align*}
\numberthis \label{e:vhatdynamics}
\vh_t &= \vh_0 e^{-\gamma t} - u \left(\int_0^t e^{-\gamma(t-s)} \nh f(\xh_0) ds \right) + \sqrt{2\gamma u} \int_0^t e^{-\gamma (t-s)} dB_s\\
\xh_t &= \xh_0 + \int_0^t \vh_s ds.
\end{align*}
\end{lemma}
\begin{Proof} Note that they have the right initial values, by setting $t=0$. By taking derivatives, one can also verify that they satisfy the differential equation \eqref{e:discretelangevindiffusionwithnoise}.
\end{Proof}

\subsection{Discretization Analysis}
In Theorem \ref{l:stochasticerrordecomposition}, we will bound the discretization error between the discrete process without noise in the gradients \eqref{e:discretelangevindiffusion} and the discrete process \eqref{e:discretelangevindiffusionwithnoise} starting from the same initial distribution.
%In particular, we bound $W_2^2 (\Phit_\delta p_0,\Phih_\delta p_0 )$. This will be sufficient to get the convergence rate stated in Theorem \ref{t:stochasticconvergence}.
\begin{lemma}\label{l:stochasticerrordecomposition}
Let $q_0$ be some initial distribution. Let $\Phit_\delta$ and $\Phih_\delta$ be as defined in \eqref{d:phi} corresponding to the discrete time process without noisy gradients and discrete-time process with noisy gradients respectively. For any $1>\delta>0$, 
$$W_2^2(\Phih_\delta q_0,q^*) = W_2^2(\Phit_\delta q_0, q^*) + \frac{5\delta^2d\sigma^2}{L^2}.$$
\end{lemma}
\begin{Proof}
Taking the difference of the dynamics in \eqref{e:vtildedynamics} and \eqref{e:vhatdynamics}, and using the definition of $\nh f(x)$. We get that 
\begin{align*}
\numberthis \label{e:vhatisvtildeplusxi}
\vh_\delta &= \vt_\delta + u\lrp{\int_0^\delta e^{-\gamma(s-\delta)} ds} \xi\\
\xh_\delta &= \xt_\delta + u\lrp{\int_0^\delta \lrp{\int_0^r e^{-\gamma(s-r)} ds} dr} \xi
\end{align*}
where $\xi$ is a zero-mean random variance with variance bounded by $\sigma^2 d$ and is independent of the Brownian motion. Let $\Gamma_1$ be the set of all couplings between $\Phit_\delta q_0$ and $q^*$ and let $\Gamma_2$ be the set of all couplings between $\Phih_\delta q_0$ and $q^*$. Let $\gamma_1(\theta,\psi)\in \Gamma_1$ be the optimal coupling between $\Phit_\delta q_0$ and $q^*$, i.e.
$$\Ep{(\theta,\psi)\sim \gamma_1}{\|\theta-\psi\|_2^2} = W_2^2 (\Phit_\delta q_0,q^*).$$
Let $\lrp{\cvec{\xt}{\wt},\cvec{x}{w}}\sim \gamma_1$. By the definition of $\gamma_1$ we have the marginal distribution of $\cvec{\xt}{\wt}\sim \Phit_\delta q_0$. Finally let us define the random variables
$$\cvec{\xh}{\wh} \triangleq \cvec{\xt}{\wt} + u\cvec{\lrp{\int_0^\delta \lrp{\int_0^r e^{-\gamma(s-r)} ds} dr}\xi}{\lrp{\int_0^\delta \lrp{\int_0^r e^{-\gamma(s-r)} ds} dr + \int_0^\delta e^{-\gamma(s-\delta)} ds} \xi}.$$

By \eqref{e:vhatisvtildeplusxi}, it follows that $\cvec{\xh}{\wh}\sim \Phih_\delta p_0$. Thus $\lrp{\cvec{\xh}{\wh},\cvec{x}{w}}$ defines a valid coupling between $\Phih_t q_0$ and $q^*$. Let us now analyze the distance between $q^*$ and $\nh_{\delta} q_0$,
\begin{align*}
& W_2^2(\Phih_\delta q_0, q^*)\\
& \overset{(i)}{\leq}  \Ep{\gamma_1}{\lrn{\cvec{\xt}{\vt} + u\cvec{\lrp{\int_0^\delta \lrp{\int_0^r e^{-\gamma(s-r)} ds} dr}\xi}{\lrp{\int_0^\delta \lrp{\int_0^r e^{-\gamma(s-r)} ds} dr + \int_0^\delta e^{-\gamma(s-\delta)} ds} \xi} - \cvec{x}{v}}_2^2}\\
& \overset{(ii)}{=} \Ep{\gamma_1}{\lrn{\cvec{\xt}{\vt}- \cvec{x}{v}}_2^2} + u\cdot\Ep{\gamma_1}{\lrn{\cvec{\lrp{\int_0^\delta \lrp{\int_0^r e^{-\gamma(s-r)} ds} dr}\xi}{\lrp{\int_0^\delta \lrp{\int_0^r e^{-\gamma(s-r)} ds} dr + \int_0^\delta e^{-\gamma(s-\delta)} ds} \xi}}_2^2}\\
& \overset{(iii)}{\leq}  \Ep{\gamma_1}{\lrn{\cvec{\xt}{\vt}- \cvec{x}{v}}_2^2} + 4u^2\lrp{\lrp{\int_0^\delta \lrp{\int_0^r e^{-\gamma(s-r)} ds} dr}^2 + \lrp{\int_0^\delta e^{-\gamma(s-\delta)} ds}^2}d\sigma^2\\
& \overset{(iv)}{\leq}  \Ep{\gamma_1}{\lrn{\cvec{\xt}{\vt}- \cvec{x}{v}}_2^2} + 4u^2\lrp{\frac{\delta^4}{4} + \delta^2}d\sigma^2\\
& \overset{(v)}{\leq} W_2^2(\Phit_t q_0, q^*) + 5u^2\delta^2d\sigma^2,
\end{align*}
where $(i)$ is by definition of $W_2$, $(ii)$ is by independence and unbiasedness of $\xi$, $(iii)$  is by Young's inequality and because $\E{\|\xi\|_2^2}\leq d\sigma^2$, $(iv)$ uses the upper bound $e^{-\gamma(s-r)}\leq 1$ and $e^{-\gamma(s-t)}\leq 1$, and finally $(v)$ is by definition of $\gamma_1$ being the optimal coupling and the fact that $\delta\leq 1$. The choice of $u=1/L$ yields the claim.

\end{Proof}
Given the bound on the discretization error between the discrete processes with and without the stochastic gradient we are now ready to prove Theorem \ref{t:stochasticconvergence}.
\begin{Proof}[Proof of Theorem \ref{t:stochasticconvergence}] 
From Corollary \ref{c:exactconvergence}, we have that for any $i \in \{1,\ldots,n \}$
$$W_2(\Phi_\d q^{(i)},q^*)\leq e^{- \delta/2\kappa}W_2(q^{(i)}, q^*).$$
By the discretization error bound in Theorem \ref{t:discretizationerror} and the sandwich inequality \eqref{e:pqsandwich}, we get
$$W_2(\Phi_\d q^{(i)}, \Phit_\d q^{(i)})\leq 2W_2(\Phi_\d p^{(i)}, \Phit_\d p^{(i)})\leq \d^2 \sqrt{\frac{8\ke}{5}}.$$
By the triangle inequality for $W_2$, 
\begin{align*}
W_2(\Phit_\d q^{(i)}, q^*) & \le W_2(\Phi_\d q^{(i)}, \Phit_\d q^{(i)}) + W_2(\Phi_\d q^{(i)},q^*)\\
& \overset{(i)}{\le} \d^2 \sqrt{\frac{8\ke}{5}} + e^{-\delta/2\kappa}W_2(q^{(i)}, q^*) \\
%& \overset{(ii)}{\leq} \d^2 \sqrt{\frac{208d}{5m}} + \left(1-\frac{\delta}{4\kappa}\right)W_2(q^{(i)}, q^*)
\end{align*}
Combining this with the discretization error bound established in Lemma \ref{l:stochasticerrordecomposition} we have,
\begin{align*}
W_2^2(\Phih_t q^{(i)},q^*) \le \left(e^{-\delta/2\kappa}W_2(q^{(i)},q^*) + \delta^2 \sqrt{\frac{8\ke}{5}} \right)^2 + \frac{5\delta^2d\sigma^2}{L^2}.
\end{align*}
By invoking Lemma \ref{l:dalalyanuseful} we can bound the value of this recursive sequence by,
\begin{align*}
W_2(q^{(n)},q^*) & \le e^{-n\delta/2\kappa} W_2(q^{(0)},q^*) + \frac{\delta^2}{1-e^{-\delta/2\kappa}}\sqrt{\frac{8\ke}{5}} \\ & \qquad \qquad \qquad \qquad \qquad \qquad \qquad \qquad + \frac{5\delta^2d\sigma^2}{L^2\left(\delta^2 \sqrt{\frac{8\ke}{5}} + \sqrt{1-e^{-\delta/\kappa}}\sqrt{\frac{5\delta^2d\sigma^2}{L^2}}\right)}.
\end{align*}
By using the sandwich inequality (Lemma \ref{l:sandwich}) we get,
\begin{align*}
W_2(p^{(n)},p^*) & \le \underbrace{4e^{-n\delta/2\kappa} W_2(p^{(0)},p^*)}_{T_1} + \underbrace{\frac{4\delta^2}{1-e^{-\delta/2\kappa}}\sqrt{\frac{8\ke}{5}}}_{T_2} \\ & \qquad \qquad \qquad \qquad \qquad \qquad \qquad \qquad + \underbrace{\frac{20\delta^2d\sigma^2}{L^2\left(\delta^2 \sqrt{\frac{8\ke}{5}} + \sqrt{1-e^{-\delta/\kappa}}\sqrt{\frac{5\delta^2d\sigma^2}{L^2}}\right)}}_{T_3}.
\end{align*}
We will now control each of these terms at a level $\varepsilon/3$. By Lemma \ref{lem:initialdistancebound} we know $W_2^2(p^{(0)},p^*) \le 3\left( \frac{d}{m} + \mathcal{D}^2 \right)$. So the choice,
\begin{align*}
n \le \frac{\kappa}{\delta}\log\left( \frac{36 \left(\frac{d}{m} + \mathcal{D}^2 \right)}{\varepsilon}\right)
\end{align*}
ensures that $T_1$ is controlled below the level $\varepsilon/3$. Note that $1-e^{-\delta/2\kappa} \ge \delta/4\kappa$ as $\delta/\kappa <1$. So the choice $\delta <\varepsilon \kappa^{-1}\sqrt{5/479232(d/m+\mathcal{D}^2)} \le \varepsilon\kappa^{-1}\sqrt{5/18432 \ke}$ (by upper bound on $\ke$ in Lemma \ref{l:kineticenergyisbounded}) ensures,
\begin{align*}
T_2  \le \frac{16\delta^2 \kappa}{\delta}\sqrt{\frac{8\ke}{5}} \le \frac{\varepsilon}{3}.
\end{align*}
Finally $\delta \le \varepsilon^2\kappa^{-1}L^2/1440d\sigma^2$ ensures $T_3$ is bounded,
\begin{align*}
T_3 & = \frac{20\delta^2d\sigma^2}{L^2\left(\delta^2 \sqrt{\frac{8\ke}{5}} + \sqrt{1-e^{-\delta/\kappa}}\sqrt{\frac{5\delta^2d\sigma^2}{L^2}}\right)} \\
& \le \frac{20\delta^2d\sigma^2}{L^2\left(\delta^2 \sqrt{\frac{8\ke}{5}} + \sqrt{\frac{5\delta^3d\sigma^2}{2L^2\kappa}}\right)} \\
& \le \frac{20\delta^2d\sigma^2}{L^2\sqrt{\frac{5\delta^3d\sigma^2}{2L^2\kappa}}} \le \frac{\varepsilon}{3}.
\end{align*}
This establishes our claim.
\end{Proof}

\section{Technical Results}
%This is a standard lemma used in the analysis of Gradient descent for smooth strong convex functions.
%\begin{lemma}[Adapted from Lemma 2.2 in \citet{GradientDes}]\label{t:graddescentlem} When $f: \mathbb{R}^d \mapsto \mathbb{R}$ is
%\begin{enumerate}
%\item twice continuously differentiable
%\item strong convex with constant $m>0$
%\item strongly smooth with constant $L>0$
%\end{enumerate}
%then we have for any $x,y$ in $\mathbb{R}^d$ ,
%\begin{align*}
%\lVert x-u\nabla f(x) - (y-u \nabla f(y)) \rVert_2 \le L_G \lVert x - y \rVert_2 
%\end{align*}
%with $L_G \le \max\{\lvert 1-uL \rvert,\lvert 1-um \rvert\}$.
%\end{lemma}
%Selecting $u=1/L$ suggests that the contraction rate $L_G \le 1-m/L$.
%We state this result from \citet{raginsky2017non} to control the variance of the invariant distribution.
%\begin{lemma}[Proposition 3.4 in \citep{raginsky2017non}] \label{t:xvariance}For any $\beta \ge 2/m$,
%\begin{align*}
%\mathbb{E}_{p^*}\left[ f(x) - f(x^*) \right] \le \frac{d}{2\beta}\log\left(\frac{eL}{m}\right).
%\end{align*}
%\end{lemma}
We state this Theorem from \citep{durmus} used in the proof of Lemma \ref{l:kineticenergyisbounded}.
\begin{theorem}[Theorem 1 in \citep{durmus}] \label{t:xvariance} For all $t\ge 0$ and $x\in \mathbb{R}^d$,
\begin{align*}
\mathbb{E}_{p^*}\left[ \lVert x - x^* \rVert_2^2 \right] \le \frac{d}{m}.
\end{align*}
\end{theorem}

The following lemma is a standard result in linear algebra regarding the determinant of a block matrix. We apply this result in the proof of Theorem \ref{t:contraction}. 
\begin{lemma}[Theorem 3 in \citep{dets}] \label{t:blockmatrix} If $A,B,C$ and $D$ are square matrices of dimension $d$, and $C$ and $D$ commute, then we have 
\begin{align*}
\det\left(\begin{bmatrix}
  A & B \\
  C & D
 \end{bmatrix} \right) = \det(AD - BC).
\end{align*}
\end{lemma}

We finally present a useful lemma from \citep{dalalyan2017user} that we will use in the proof of Theorem \ref{t:stochasticconvergence}.

\begin{lemma}[Lemma 7 in \citep{dalalyan2017user}] \label{l:dalalyanuseful} Let $A$, $B$ and $C$ be given non-negative numbers such that $A \in \{0,1\}$. Assume that the sequence of non-negative numbers $\{x_k\}_{k \in \mathbb{N}}$ satisfies the recursive inequality
\begin{align*}
x_{k+1}^2 \le \left[(A)x_k + C\right]^2 + B^2
\end{align*}
for every integer $k \ge 0$. Then
\begin{align}
x_k \le A^{k}x_0 + \frac{C}{1-A} + \frac{B^2}{C + \sqrt{(1-A^2)}B}
\end{align}
for all integers $k \ge 0$.

\end{lemma}
%\section{Alternative Proofs}
%\input{alternative}
\end{document}